\documentclass[conference]{IEEETran}
\usepackage[utf8]{inputenc} 
\usepackage[T1]{fontenc}    
\usepackage{hyperref}       
\usepackage{url}            
\usepackage{booktabs}       
\usepackage{amsfonts}       
\usepackage{nicefrac}       
\usepackage{microtype}      
\usepackage{xcolor}         
\usepackage{amsmath,amssymb}
\usepackage{wrapfig}
\usepackage{stfloats}
\usepackage{graphicx}
\usepackage{float}
\usepackage{caption}
\usepackage{subcaption}
\usepackage{bm}
\usepackage{listings}
\usepackage{multicol}
\usepackage{multirow} 

\NewDocumentCommand{\codeword}{v}{%
\texttt{\textcolor{blue}{#1}}%
}

\title{A Gated Residual \textbf{K}olmogorov-\textbf{A}rnold Networks for \textbf{M}ixtures \textbf{o}f \textbf{E}xperts}

\author{%
  \IEEEauthorblockN{%
    Hugo Inzirillo\IEEEauthorrefmark{1}\textsuperscript{\textsection} and
    Rémi Genet\IEEEauthorrefmark{2}\textsuperscript{\textsection}
  }%
  
  \IEEEauthorblockA{\IEEEauthorrefmark{1} CREST-ENSAE, Institut Polytechnique de Paris}%
    \IEEEauthorblockA{\IEEEauthorrefmark{2} DRM, Université Paris Dauphine - PSL}%
}

\begin{document}

\thispagestyle{plain}
\pagestyle{plain}
\maketitle
\begingroup\renewcommand\thefootnote{\textsection}
\footnotetext{These authors contributed equally.}
\endgroup

\begin{abstract}
This paper introduces KAMoE, a novel Mixture of Experts (MoE) framework based on Gated Residual Kolmogorov-Arnold Networks (GRKAN). We propose GRKAN as an alternative to the traditional gating function, aiming to enhance efficiency and interpretability in MoE modeling. Through extensive experiments on digital asset markets and real estate valuation, we demonstrate that KAMoE consistently outperforms traditional MoE architectures across various tasks and model types. Our results show that GRKAN exhibits superior performance compared to standard Gating Residual Networks, particularly in LSTM-based models for sequential tasks. We also provide insights into the trade-offs between model complexity and performance gains in MoE and KAMoE architectures. 

\begin{figure*}[b!]
    \centering
    \includegraphics[scale=0.4]{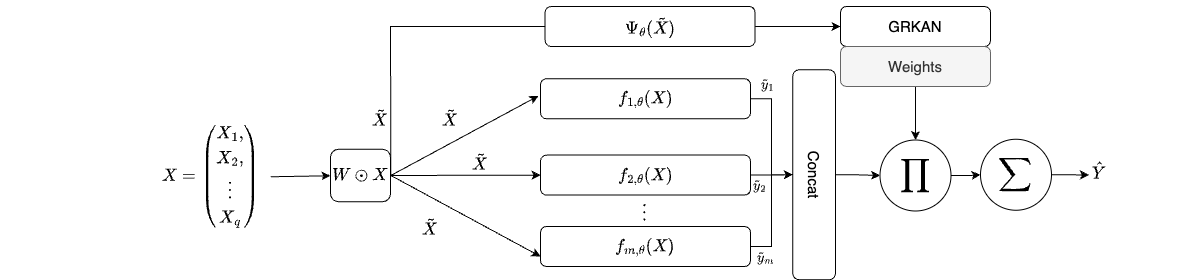}
    \caption{KAMoE} 
    \label{fig:KAMoE}
\end{figure*}

\end{abstract}

\section{Introduction}
Introduced by \cite{jacobs1991adaptive,jordan1994hierarchical}, the Mixture of Experts (MoE) is one of the most popular machine learning techniques. It involves the combination of multiple neural networks, where each network called “expert”, focusing on specific parts of the input. A gating network is used to assign different weights to these "experts". This gating mechanism is deciding which expert should be most trusted for a given input. This approach helps in improving the overall performance of the system by leveraging the strengths of individual neural networks specialized in different tasks. The MoE has since been applied to various domains such natural language processing \cite{shazeer2017outrageously,du2022glam}, computer vision \cite{enzweiler2011multilevel,ahmed2016network,riquelme2021scaling}, and reinforcement learning \cite{aljundi2017expert}. Time series data can exhibit complex, non-linear patterns and regime shifts \cite{inzirillo2024deep} . The MoE's ability to combine multiple specialized expert networks can help capture these complex temporal dynamics observed in financial time series \cite{ebrahimpour2011mixture}. Despite the increase in the availability of datasets for time series forecasting, it remains a very complex task. Powerful models have been proposed through the years \cite{salinas2020deepar,oreshkin2019n}.  Some research proposed dynamic systems to model different existing states within a time series \cite{rangapuram2018deep,li2021learning,inzirillo2024deep}. In this paper, we explore the Mixture of Experts (MoE) using a novel neural networks architecture embedded in a gating mechanism. This new gating mechanism is built over Kolmogorov-Arnold Networks \cite{liu2024kan}, allowing for a more accurate estimation of weights for each neural network denoted $f_{i,\theta} \text{ and } i \in [0;n]$.

\medskip

Recently,  Liu et al. \cite{liu2024kan} released the Kolmogorov-Arnold Networks (KANs) a promising architecture introduced as an alternative for MLPs. To dive into the architecture of the model, we redirect the reader to \cite{liu2024kan}. In our previous work we have adapted KANs for time series forecasting \cite{genet2024tkan}, other researchers also proposed an adaptation for time series \cite{xu2024kolmogorov}. According to \cite{vaca2024kolmogorov}, KANs outperform conventional MLPs in real-world forecasting tasks. In a previous work we also proposed a temporal transformer architecture using TKANs \cite{genet2024temporal}. Other extensions using wavelets \cite{bozorgasl2024wav} have been proposed in the meantime. TKANs aim to develop a framework with two key functionalities: handling sequential data and managing memory. To do so, in our previous work \cite{genet2024tkan} we enriched the initial model by adding a recurring layer of KANs to introduce RKANs. Temporal Kolmogorov-Arnold Networks (TKANs) is an upgraded version of LSTM \cite{hochreiter1997long} using Kolmogorov-Arnold Networks (KANs). They rely on RKANs for the management of short-term memory as well as a cell state; for further details, we refer the reader to \cite{genet2024tkan}.

\medskip

In this paper, we rely on the gated residual kolmogorov-arnold networks \cite{inzirillo2024sigkan} to estimate weight for each expert of the framework. All the experiments and codes are available at \href{https://github.com/remigenet/kamoe}{KAMoE repository} and the package can be installed using the following command: \codeword{pip install kamoe}. The data are accessible if a reader wishes to reproduce the experiments inside the github provided above.

\section{Framework}
In this paper, we introduce a new framework called \textbf{“KAMoE”} Figure \ref{fig:KAMoE}, based on Gated Residual Kolmogorov-Arnold Networks (GRKAN) introduced in our previous work \cite{inzirillo2024sigkan}. Each expert in charge of focusing on a specific part of the sequence requires optimal weighting. This weighting is achieved through a gating mechanism. A gated network is used to assign different weights to these “experts”. This gating mechanism is deciding which expert should be most trusted for a given input. This approach helps in improving the overall performance of the system by leveraging the strengths of individual neural networks specialized in different tasks. To assign these weights we will use the GRKAN.

\subsection{Inputs}
The \textbf{KAMoE} is able to manage sequential and non-sequential data. It allows to leverage the experts according to specific tasks. We define $X_i=(X_{i,1},X_{i,2},...,X_{i,q})$ the vector of inputs where $X_{i,j} \in \mathbb{R}^{s}, \quad j \in [1,2,...,q]$ and $s$ denotes the length of the input sequence. The particularity in the framework is that the input will be considered as learnable inputs denoted $\Tilde{x}$ such

\begin{equation}
    \Tilde{x_i} = w_{\Tilde{x_i}} \odot X_i,
    \label{eq:transformed_inputs}
\end{equation}

where $W_{\Tilde{x_i}} \in \mathbb{R}^{dim(X)}$. The dimension of X noted $dim(X)$ may vary according to the inputs. In the case of sequential data we will have a 2D weight matrix. This transformed input with learnable parameter will be the input of the gated mechanism as well as all the experts defined in a latter section.

\subsection{Gated Mechanism}
Establishing relationship between temporal data is a key issue. Gated Residual Networks (GRNs) offer an efficient and flexible way of modelling complex relationships in data. They allow to control the flow of information and facilitate the learning tasks. They are particularly useful in areas where nonlinear interactions and long-term dependencies are crucial. In our model we use the Gated Residual Kolmogorov-Arnold Networks (GRKAN) inspired from the GRN proposed by \cite{lim2021temporal}, we kept the same architecture. We propose a new approach using two KAN Linear layers \cite{liu2024kan} , to control the information flow while bringing more interpretability. Using GRKAN there is no more need for context which is contains in path signature, however, an additional linear layer is required to match the signature transform and the ouput of the gating mechanism.
\begin{figure}
    \centering
    \includegraphics[width=0.65\linewidth]{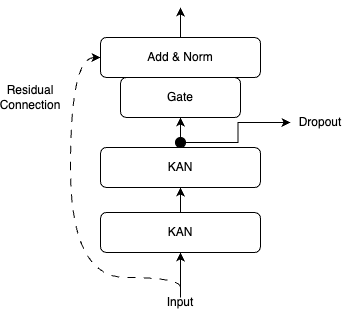}
    \caption{Gated Residual KAN (GRKAN)} 
    \label{fig:GRKAN}
\end{figure}
\begin{align}
\text{GRKAN}_\omega\left(x \right) &=\text{LayerNorm}\left(x + \text{GLU}_\omega(\eta_1) \right), \\
    \eta_1 &= \text{KAN}(\varphi_{\eta_1}(.),\eta_2),  \label{eq:grn_step}\\
    \eta_2 &= \text{KAN}(\varphi_{\eta_2}(.),x). & 
\label{eq:grkan}    
\end{align}
In this context, we used activation functions for KAN layers denoted, $\varphi_{\eta_1}(.)$ and $\varphi_{\eta_2}(.)$, $\text{SiLU}$ \cite{elfwing2018sigmoid} and ELU \cite{clevert2015fast}, respectively, while $\eta_1 \in \mathbb{R}^{d_{model}}$ and $\eta_2 \in \mathbb{R}^{d_{model}}$ represent intermediate layers. The standard layer normalization $\text{LayerNorm}$ is that described in \cite{ba2016layer}, and $\omega$ is an index used to indicate weight sharing. When the expression $\text{KAN}(x)$ is largely positive, ELU activation works as an identity function. On the other hand, when this expression is largely negative, ELU activation produces a constant output, thus behaving like a linear layer. We use Gated Linear Units (GLUs) \cite{dauphin2017language} to provide the flexibility to suppress any parts of the architecture that are not required for a given dataset. Letting $\gamma \in \mathbb{R}^{d_{model}}$ be the input, the GLU then takes the form:
\begin{align}
 \text{GLU}_\omega(\gamma) & =  \sigma(W_{4, \omega}~\gamma + b_{4, \omega}) \odot (W_{5, \omega}~\gamma + b_{5, \omega} ),
\label{eq:component_gate}
\end{align}
where $\sigma(.)$ is the sigmoid activation function, $W_{(.)} \in \mathbb{R}^{d_{model}\times d_{model}}, b_{(.)} \in \mathbb{R}^{d_{model}}$ are the weights and biases, $\odot$ is the element-wise Hadamard product, and $d_{model}$ is the hidden state size. GLU allows to control the extent to which the GRN contributes to the original input $x$ -- potentially skipping over the layer entirely if necessary as the GLU outputs could be all close to 0 in order to suppress the nonlinear contribution.

\subsection{Experts}
We denote $f_{k,\theta_k}(.)$ a highly parametrized function which describes the succession of operations of neural networks such GRU, LSTM and TKAN. Let us denote $m$ the total number of experts, $k\in \{1,...,m\}$. The role of the expert in the framework is the learn from patterns within the sequence in the dataset. Each output of the expert is denoted $\Tilde{y}_k$ and obtained simultaneously from the transformed input defined Eq. \eqref{eq:transformed_inputs}

\begin{equation}
    \Tilde{y}_{i,k} = f_{k,\theta_k}(\Tilde{x_i}).
\end{equation}

All the estimates per experts will be  stacked in $\Tilde{y}_i$

\begin{equation}
    \Tilde{y}_i := \text{Concat}[\Tilde{y}_{i,1},\Tilde{y}_{i,2},...,\Tilde{y}_{i,m}].
\end{equation}

\subsection{Output}
The weight of each expert will be determined using the output of the GRKAN Eq.\eqref{eq:grkan}, for simplication we will denote the succession of the GRKAN operation $\Xi(.)$. Before the evaluation of these weights.  

\begin{equation}
    z_i = \Psi_{\psi}(\Tilde{x_i}).
\end{equation}

The output of this transformation will feed the GRKAN layer to obtain the weight. These outputs will be converted to weights,
\begin{equation}
    a_i = \sigma(\Xi(z_i))
\end{equation}
where $a_i\in \mathbb{R}^m$ and contains each weight for each expert and $\sigma(.)$ is a sigmoid activation function. The global output of our mixture will be obtained combining the weights and the ouput of each expert such
\begin{equation}
    \begin{split}
        \hat{y}_i  &= \sum_{k}^{m} a_{i,k} f_{k,\theta_k}(x)\\
        &= \sum_{k}^{m} a_{i,k} \Tilde{y}_{i,k}
    \end{split}
\end{equation}
where $f_{k,\theta}(.)\in \mathbb{R}^{out}$ where $out$ is the output dimension of our mixture.

\section{Learning}

As mixture of experts is a method that can be applied to any networks, we created two learning tasks in order to test it in both a sequential case with temporality and in a more standard one.

\medskip

The first learning task is inspired by the one in \cite{genet2024tkan}, \cite{genet2024temporal} and \cite{inzirillo2024sigkan} with a few modifications. The task is about predicting market trading notional multiple steps ahead on one asset using the past values of the asset but also others. We chose this task as it's a challenging one to perform due to the very noisy series but also the internal patterns in the traded volume like strong auto-correlation and seasonality.

\medskip

We differ from the previous papers by reducing the number of input assets, which are BTC, ETH, ETC, XRP, BCH, LTC, and not only predicting BTC but instead training models and analyzing results for each of them. Everything else is similar to the previous papers; the data are from Binance and cover the period from January 1, 2020 to December 31, 2022. The data are preprocessed by dividing the values in the series by the moving median of the last two weeks and applying a MinMax Scaling between 0 and 1.

\medskip

The model is kept simple to study the effect of MoE and KAMoE on them. As we are on a sequential prediction task, we used the two most standard RNN layers that are Gated Recurrent Unit (GRU), and Long Short Term Memory (LSTM).
Each model is done by applying a first layer that returns the full sequence, followed by a second that only returns the last element. Finally, this output is fully connected with a linear layer that has the same number of units as there are steps to predict. 
The two hidden RNN layers are both composed of 100 units.
Finally, to test the MoE and KAMoE effect, we only apply it on the first hidden RNN layer, which is composed of 3 experts and uses a sigmoid activation on the GRKAN outputs.

\medskip

The second learning task is a more standard one, utilizing the California Housing dataset, which is commonly used in ML teaching and research. We obtained this dataset from scikit-learn. This dataset, derived from the 1990 U.S. census, focuses on housing in California and contains 20,640 instances with 8 numeric features. These features include median income, house age, average number of rooms, population, and geographical coordinates (latitude and longitude).
The target variable for this task is the median house value for California districts, expressed in hundreds of thousands of dollars. This presents a regression problem, where the goal is to predict the median house value based on the given features.

\medskip

For preprocessing the California Housing data, we applied StandardScaler fitted on the training set. The StandardScaler standardizes features by removing the mean and scaling to unit variance, which is indeed a good and standard approach for many machine learning algorithms, especially when dealing with features on different scales. This preprocessing step helps to ensure that all features contribute equally to the model and can improve the convergence of many machine learning algorithms.

\medskip

We chose this dataset for several reasons:
\begin{enumerate}
\item It is not a sequential one so we can test the usefulness of our KAMoE and MoE in general not on time series
\item It represents a more traditional machine learning task, contrasting with our first, time-series-based task.
\item Its medium size and real-world nature provide a good balance between complexity and practicality.
\item As a widely-used benchmark, it facilitates comparison with other machine learning approaches in the literature.
\end{enumerate}
By applying our mixture of experts method to these two distinct learning tasks, we aim to demonstrate its versatility across different types of data structures and problem domains.

\medskip

For this task we compare the KAMoE effect on non-sequential model, we thus used very simple model which are an MLP and a KAN model. The MLP is composed of successive layers with ReLU activation, while not applying other activation to the KAN. 
We wanted to take the opportunity of this task to see the effect of the number of layers and hidden size for both, as well as being able to analyze whether the addition of the mixture framework does not increase results just by increasing the number of units.
Finally we also compared the results with the one obtained with standard machine learning model using the one from scikit-learn as well as XGBoost.

\subsection{Task 1: Results}

The results from the first task reveal that Mixture of Experts (MoE) is not a universally applicable solution for improving simple Recurrent Neural Networks (RNNs). Instead, its effectiveness varies depending on the specific context and model architecture. This variability is evident in two key aspects:

\begin{enumerate}
    \item \textbf{Differential impact on GRU vs. LSTM:}
    The MoE framework shows markedly different effects when applied to Gated Recurrent Units (GRUs) compared to Long Short-Term Memory (LSTM) networks. While it generally yields significant improvements for LSTM models, the results for GRU-based models are more nuanced and context-dependent.

    \item \textbf{Asset-specific performance:}
    The effectiveness of MoE varies across different cryptocurrencies. For instance, while it demonstrates notable improvements in Bitcoin (BTC) predictions, its performance on other assets is less consistent and more debatable.
\end{enumerate}

These findings underscore that MoE is a sophisticated framework that should be applied judiciously, considering the specific characteristics of the problem at hand and the base model architecture.

A significant outcome of this study is the performance of our novel Kolmogorov-Arnold gating residual network (GRKAN), which replaces the standard Gating Residual Network (GRN) in the MoE framework. The GRKAN consistently outperforms the standard GRN across most scenarios, demonstrating the efficacy of incorporating Kolmogorov-Arnold principles into the gating mechanism of the mixture.

Examining the global average differences versus standard models, we observe:

\begin{itemize}
    \item For GRU-based models:
    \begin{itemize}
        \item MoE shows a slight average decrease in performance (-0.007)
        \item KAMoE demonstrates a small average improvement (0.005)
    \end{itemize}
    \item For LSTM-based models:
    \begin{itemize}
        \item Both MoE and KAMoE show more substantial improvements (0.015 and 0.017 respectively)
    \end{itemize}
\end{itemize}

These results highlight that while MoE and KAMoE architectures can offer improvements, their effectiveness is not uniform across all scenarios. The KAMoE variant, in particular, shows promise by consistently outperforming or matching the standard MoE in most cases.

\subsection{Task 2: Results}
\begin{table*}[t]
\centering
\caption{KANLinear: Mean R2 Values}
\newcommand{\bestrow}[1]{\textbf{#1}}
\newcommand{\bestmode}[1]{\underline{#1}}
\begin{tabular}{|c|cccc|cccc|cccc|}
\hline
\multirow{2}{*}{\shortstack{Hidden\\Units}} & \multicolumn{4}{c|}{KAMoE} & \multicolumn{4}{c|}{MoE} & \multicolumn{4}{c|}{Standard} \\
 & \multicolumn{12}{c|}{Number of Layers} \\
 & 1 & 2 & 3 & 4 & 1 & 2 & 3 & 4 & 1 & 2 & 3 & 4 \\
\hline
5 & \bestmode{0.767} & 0.762 & \bestmode{0.766} & \bestmode{0.755} & 0.758 & \bestrow{\bestmode{0.769}} & 0.757 & 0.751 & 0.621 & 0.691 & 0.688 & 0.671 \\
10 & 0.765 & \bestrow{\bestmode{0.777}} & \bestmode{0.760} & \bestmode{0.753} & \bestmode{0.768} & 0.771 & 0.758 & 0.751 & 0.621 & 0.705 & 0.697 & 0.683 \\
25 & 0.768 & \bestmode{0.770} & \bestmode{0.753} & 0.744 & \bestrow{\bestmode{0.772}} & 0.767 & \bestmode{0.750} & 0.736 & 0.621 & 0.699 & 0.694 & 0.682 \\
50 & \bestrow{\bestmode{0.776}} & \bestmode{0.765} & \bestmode{0.751} & 0.745 & 0.775 & 0.761 & \bestmode{0.745} & 0.747 & 0.622 & 0.699 & 0.686 & 0.684 \\
100 & \bestrow{\bestmode{0.783}} & \bestmode{0.765} & \bestmode{0.760} & 0.756 & 0.780 & 0.762 & 0.748 & \bestmode{0.758} & 0.621 & 0.693 & 0.684 & 0.685 \\
200 & 0.783 & 0.762 & 0.765 & \bestmode{0.770} & \bestrow{\bestmode{0.784}} & \bestmode{0.763} & \bestmode{0.768} & \bestmode{0.765} & 0.622 & 0.693 & 0.689 & 0.698 \\
400 & 0.781 & \bestmode{0.748} & \bestmode{0.768} & \bestmode{0.772} & \bestrow{\bestmode{0.785}} & 0.746 & \bestmode{0.768} & 0.771 & 0.622 & 0.689 & 0.699 & 0.705 \\
800 & 0.782 & 0.741 & 0.765 & \bestmode{0.773} & \bestrow{\bestmode{0.789}} & \bestmode{0.742} & \bestmode{0.769} & \bestmode{0.773} & 0.622 & 0.696 & 0.703 & 0.708 \\
\midrule
\multicolumn{1}{|l|}{\textbf{Avg. Diff. vs. Standard}} & 0.154145 & 0.065672 & 0.068344 & 0.068918 & 0.154954 & 0.064504 & 0.065217 & 0.067085 & - & - & - & - \\
\bottomrule
\end{tabular}
\end{table*}
The results obtained from our experiments yield several significant insights into the performance of various models on the given task.
\subsection{Comparison of KAN and MLP Models}
Our findings indicate that the KAN model, when used independently, falls short of matching the MLP (Multi-Layer Perceptron) in terms of efficiency. However, both models exhibit similar patterns when increasing the number of units or hidden layers. Specifically:
\begin{enumerate}
\item Increasing the number of units generally improves model performance, albeit at the cost of increased complexity.
\item The optimal number of hidden layers appears to be two for both models, with additional layers providing negligible benefits.
\end{enumerate}
\begin{table}[h]
\centering
\begin{tabular}{|l|c|c|}
\hline
\textbf{Model} & \textbf{R2} & \textbf{RMSE} \\
\hline
KNeighborsRegressor & 0.67001 & 0.432422 \\
Lasso & -0.000219 & 1.310696 \\
LinearRegression & 0.575788 & 0.555892 \\
RandomForest & 0.774564 & 0.295413 \\
Ridge & 0.575816 & 0.555855 \\
SVR & 0.727564 & 0.357003 \\
XGBoost & 0.537515 & 0.606044 \\
\hline
\end{tabular}
\caption{Comparison of Machine Learning Models}
\label{tab:model-comparison}
\end{table}
\subsection{Impact of Mixture of Experts (MoE) and Kernel Activation Mixture of Experts (KAMoE)}
The introduction of MoE and KAMoE architectures demonstrates substantial improvements in performance, particularly for smaller models. However, this effect diminishes for MLPs with a high number of units. In contrast, for the KAN model, the adoption of mixture architectures proves to be a game-changer, consistently enhancing performance across various configurations. Notably:
\begin{itemize}
\item For KAN models, the ideal number of layers becomes one when using mixture architectures, reducing model complexity.
\item The number of parameters increases significantly with the number of units in MoE and KAMoE models, due to our choice of using an equal number of hidden units across all layers. This approach may not be optimal and could be refined to reduce the parameter count.
\end{itemize}
\subsection{Performance of KAMoE}
One of the most intriguing findings is that KAMoE consistently outperforms MoE for both model types. While the standalone performance of KAN may be debatable, it undeniably serves as a valuable tool for enhancing model performance when strategically implemented.
\subsection{Comparison with Standard Machine Learning Models}
We also evaluated standard machine learning models for comparison:
\begin{itemize}
\item Neural network approaches generally yielded superior results compared to traditional machine learning models.
\item The Random Forest model stood out among traditional methods, achieving performance comparable to neural networks. However, it's worth noting that Random Forest models can be highly complex, potentially rivaling neural networks in this aspect.
\end{itemize}

\begin{table*}[t]
\centering
\caption{MLP: Mean R2 Values}
\newcommand{\bestrow}[1]{\textbf{#1}}
\newcommand{\bestmode}[1]{\underline{#1}}
\begin{tabular}{|c|cccc|cccc|cccc|}
\hline
\multirow{2}{*}{\shortstack{Hidden\\Units}} & \multicolumn{4}{c|}{KAMoE} & \multicolumn{4}{c|}{MoE} & \multicolumn{4}{c|}{Standard} \\
 & \multicolumn{12}{c|}{Number of Layers} \\
 & 1 & 2 & 3 & 4 & 1 & 2 & 3 & 4 & 1 & 2 & 3 & 4 \\
\hline
5 & 0.753 & 0.764 & \bestmode{0.769} & \bestrow{\bestmode{0.768}} & \bestmode{0.755} & \bestmode{0.762} & 0.767 & 0.758 & 0.695 & 0.728 & 0.726 & 0.727 \\
10 & \bestmode{0.768} & \bestmode{0.775} & 0.772 & 0.774 & 0.764 & 0.769 & \bestmode{0.776} & \bestrow{\bestmode{0.780}} & 0.733 & 0.755 & 0.759 & 0.764 \\
25 & \bestmode{0.775} & 0.777 & \bestmode{0.782} & 0.783 & \bestmode{0.775} & \bestmode{0.783} & 0.781 & 0.783 & 0.751 & 0.773 & 0.780 & \bestrow{\bestmode{0.785}} \\
50 & \bestmode{0.777} & \bestmode{0.786} & \bestmode{0.788} & 0.787 & 0.774 & 0.785 & 0.787 & \bestrow{\bestmode{0.789}} & 0.756 & 0.781 & 0.785 & 0.786 \\
100 & 0.781 & \bestmode{0.791} & \bestrow{\bestmode{0.795}} & 0.792 & \bestmode{0.782} & 0.788 & 0.792 & 0.792 & 0.762 & 0.790 & 0.794 & \bestmode{0.793} \\
200 & 0.783 & 0.792 & \bestmode{0.796} & \bestrow{\bestmode{0.797}} & \bestmode{0.784} & 0.792 & 0.794 & \bestrow{\bestmode{0.797}} & 0.764 & \bestmode{0.794} & 0.795 & 0.796 \\
400 & 0.785 & 0.796 & 0.799 & \bestmode{0.797} & \bestmode{0.786} & 0.798 & 0.798 & 0.795 & 0.765 & \bestrow{\bestmode{0.801}} & \bestmode{0.800} & \bestmode{0.797} \\
800 & \bestmode{0.788} & 0.798 & \bestmode{0.801} & 0.798 & 0.786 & 0.797 & 0.798 & 0.798 & 0.772 & \bestrow{\bestmode{0.803}} & 0.800 & \bestmode{0.800} \\
\hline
\midrule
\multicolumn{1}{|l|}{\textbf{Avg. Diff. vs. Standard}} & 0.027 & 0.007 & 0.0.008 & 0.006 & 0.026 & 0.006 & 0.007 & 0.005 & - & - & - & - \\
\bottomrule
\end{tabular}
\end{table*}

\section{Conclusion}
In this paper, we introduced KAMoE, a novel framework for Mixture of Experts (MoE) modeling based on Gated Residual Kolmogorov-Arnold Networks (GRKAN). Our comprehensive experiments across both sequential and non-sequential tasks yield several significant insights:

\begin{enumerate}
    \item \textbf{Context-Dependent Efficacy}: While MoE and KAMoE demonstrate potential for improving model performance, their effectiveness varies depending on the specific context and model architecture. This underscores the importance of judicious application of these techniques.

    \item \textbf{GRKAN Superiority}: Our proposed GRKAN consistently outperforms standard Gating Residual Networks (GRN) across most scenarios, showcasing the efficacy of incorporating Kolmogorov-Arnold principles into the gating mechanism of MoE.

    \item \textbf{Model-Specific Impact}: The impact of MoE and KAMoE varies across different model types. For instance, LSTM-based models showed more substantial improvements compared to GRU-based models when integrated with these techniques.

    \item \textbf{KAMoE Performance}: KAMoE consistently outperformed or matched standard MoE in most cases, highlighting its potential as a powerful enhancement to existing MoE architectures.

    \item \textbf{Complexity vs. Performance Trade-off}: While MoE and KAMoE architectures can offer significant performance improvements, they also increase model complexity and computational requirements. This trade-off needs careful consideration in practical applications.

    \item \textbf{KAN as a Strategic Tool}: Although standalone Kolmogorov-Arnold Networks (KAN) may not always outperform traditional architectures like MLPs, our results demonstrate that KAN can be a valuable tool when strategically implemented within larger frameworks like KAMoE.
\end{enumerate}

These findings open up new avenues for research in adaptive neural network architectures. Future work could explore optimizing the balance between model complexity and performance in KAMoE, investigating its applicability in other domains, and further refining the integration of Kolmogorov-Arnold principles in neural network design. Our work contributes to the ongoing effort to develop more efficient and adaptable machine learning models, potentially impacting a wide range of applications from time series forecasting to complex regression tasks.

\bibliographystyle{IEEEtran}
\bibliography{bib}

\appendices
\onecolumn
\section{Additional Results Task 1}
\begin{table}[H]
\centering
\caption{Mean R2 Scores}
\label{tab:mean-r2-scores}
\newcommand{\bestrow}[1]{\textbf{#1}}
\newcommand{\bestgru}[1]{\underline{#1}}
\newcommand{\bestlstm}[1]{\underline{#1}}
\resizebox{\textwidth}{!}{%
\begin{tabular}{llrrrrrr}
\toprule
Asset & Steps & GRU & MoE\_GRU & KAMoE\_GRU & LSTM & MoE\_LSTM & KAMoE\_LSTM \\
\midrule
\multirow{6}{*}{BTC} & 1 & \bestgru{0.398} & 0.391 & 0.392 & \bestrow{\bestlstm{0.399}} & 0.371 & 0.385 \\
 & 3 & \bestrow{\bestgru{0.239}} & 0.227 & 0.233 & 0.129 & 0.178 & \bestlstm{0.205} \\
 & 6 & 0.090 & 0.115 & \bestrow{\bestgru{0.155}} & -0.232 & 0.000 & \bestlstm{0.088} \\
 & 9 & -0.045 & \bestrow{\bestgru{0.073}} & 0.054 & -0.206 & \bestlstm{-0.047} & -0.107 \\
 & 12 & -0.117 & -0.044 & \bestgru{\bestrow{0.019}} & -0.449 & -0.126 & \bestlstm{-0.080} \\
 & 15 & -0.032 & \bestrow{\bestgru{0.028}} & -0.017 & -0.222 & -0.142 & \bestlstm{-0.115} \\
\midrule
\multicolumn{2}{l}{Avg. Diff. vs. Standard} & - & 0.042 & 0.051 & - & 0.136 & 0.092 \\
\midrule
\multirow{6}{*}{LTC} & 1 & 0.372 & 0.379 & \bestrow{\bestgru{0.384}} & 0.375 & 0.370 & \bestlstm{0.376} \\
 & 3 & 0.272 & 0.262 & \bestgru{\bestrow{0.276}} & 0.263 & 0.258 & \bestlstm{0.268} \\
 & 6 & \bestrow{\bestgru{0.227}} & 0.216 & 0.226 & \bestlstm{0.213} & 0.203 & 0.211 \\
 & 9 & \bestrow{\bestgru{0.211}} & 0.189 & 0.205 & \bestlstm{0.194} & 0.179 & 0.184 \\
 & 12 & \bestrow{\bestgru{0.199}} & 0.185 & 0.193 & \bestlstm{0.187} & 0.172 & 0.184 \\
 & 15 & 0.181 & \bestgru{\bestrow{0.182}} & 0.181 & \bestlstm{0.178} & 0.164 & 0.162 \\
\midrule
\multicolumn{2}{l}{Avg. Diff. vs. Standard} & - & -0.008 & 0.001 & - & -0.010 & 0.001 \\
\midrule
\multirow{6}{*}{ETC} & 1 & 0.491 & 0.491 & \bestgru{0.493} & 0.497 & \bestrow{\bestlstm{0.508}} & 0.503 \\
 & 3 & 0.445 & \bestgru{0.450} & 0.447 & 0.448 & \bestrow{\bestlstm{0.454}} & 0.448 \\
 & 6 & \bestrow{\bestgru{0.428}} & 0.422 & 0.426 & 0.421 & 0.418 & \bestlstm{0.424} \\
 & 9 & \bestrow{\bestgru{0.421}} & 0.373 & 0.420 & 0.403 & \bestlstm{0.413} & 0.410 \\
 & 12 & \bestrow{\bestgru{0.410}} & 0.363 & 0.402 & \bestlstm{0.403} & 0.359 & 0.391 \\
 & 15 & 0.404 & 0.356 & \bestgru{\bestrow{0.405}} & 0.391 & 0.378 & \bestlstm{0.394} \\
\midrule
\multicolumn{2}{l}{Avg. Diff. vs. Standard} & - & -0.0240 & -0.001 & - & -0.006 & 0.004 \\
\midrule
\multirow{6}{*}{ETH} & 1 & 0.315 & \bestgru{0.318} & 0.317 & 0.322 & 0.322 & \bestrow{\bestlstm{0.325}} \\
 & 3 & \bestgru{0.226} & 0.220 & 0.221 & \bestrow{\bestlstm{0.230}} & 0.221 & 0.214 \\
 & 6 & \bestrow{\bestgru{0.169}} & 0.132 & 0.160 & 0.149 & \bestlstm{0.161} & 0.149 \\
 & 9 & \bestrow{\bestgru{0.140}} & 0.124 & 0.134 & 0.125 & 0.131 & \bestlstm{0.140} \\
 & 12 & 0.129 & 0.124 & \bestrow{\bestgru{0.137}} & 0.119 & 0.116 & \bestlstm{0.127} \\
 & 15 & \bestrow{\bestgru{0.129}} & 0.114 & 0.094 & \bestlstm{0.128} & 0.111 & 0.115 \\
\midrule
\multicolumn{2}{l}{Avg. Diff. vs. Standard} & - & -0.013 & -0.007 & - & -0.002 & +0.000 \\
\midrule
\multirow{6}{*}{XRP} & 1 & 0.423 & \bestgru{0.433} & 0.431 & 0.437 & 0.432 & \bestrow{\bestlstm{0.438}} \\
 & 3 & \bestgru{0.337} & 0.322 & 0.336 & 0.323 & 0.313 & \bestrow{\bestlstm{0.342}} \\
 & 6 & 0.280 & 0.230 & \bestgru{\bestrow{0.284}} & 0.264 & 0.251 & \bestlstm{0.260} \\
 & 9 & \bestrow{\bestgru{0.265}} & 0.207 & 0.238 & \bestlstm{0.254} & 0.224 & 0.239 \\
 & 12 & \bestrow{\bestgru{0.258}} & 0.208 & 0.224 & \bestlstm{0.238} & 0.213 & 0.224 \\
 & 15 & \bestrow{\bestgru{0.238}} & 0.182 & 0.210 & \bestlstm{0.235} & 0.184 & 0.190 \\
\midrule
\multicolumn{2}{l}{Avg. Diff. vs. Standard} & - & -0.036 & -0.013 & - & -0.022 & 0.0013 \\
\midrule
\multirow{6}{*}{BCH} & 1 & 0.265 & \bestgru{0.270} & 0.268 & 0.271 & 0.271 & \bestrow{\bestlstm{0.273}} \\
 & 3 & 0.180 & \bestrow{\bestgru{0.184}} & 0.181 & 0.180 & \bestlstm{0.182} & 0.180 \\
 & 6 & \bestrow{\bestgru{0.139}} & 0.134 & \bestgru{0.139} & \bestlstm{0.138} & 0.135 & 0.137 \\
 & 9 & \bestrow{\bestgru{0.113}} & 0.107 & 0.104 & \bestlstm{0.111} & 0.110 & \bestlstm{0.111} \\
 & 12 & \bestrow{\bestgru{0.101}} & 0.095 & 0.094 & 0.095 & 0.084 & \bestlstm{0.096} \\
 & 15 & \bestrow{\bestgru{0.096}} & 0.075 & 0.088 & \bestrow{\bestlstm{0.096}} & 0.077 & 0.088 \\
\midrule
\multicolumn{2}{l}{Avg. Diff. vs. Standard} & - & -0.005 & -0.003 & - & -0.005 & 0.002 \\
\midrule
\multicolumn{2}{l}{\textbf{Global Avg. Diff. vs. Standard}} & - & -0.007 & 0.005 & - & 0.015 & 0.017 \\
\bottomrule
\end{tabular}%
}
\end{table}

\begin{table}[H]
\centering
\caption{Standard Deviation of R2 Scores}
\label{tab:std-r2-scores}
\resizebox{\textwidth}{!}{%
\begin{tabular}{llrrrrrr}
\toprule
Asset & Steps & GRU & MoE\_GRU & KAMoE\_GRU & LSTM & MoE\_LSTM & KAMoE\_LSTM \\
\midrule
\multirow{6}{*}{BTC} & 1 & 0.004 & 0.003 & 0.005 & 0.007 & 0.004 & 0.012 \\
 & 3 & 0.007 & 0.005 & 0.009 & 0.051 & 0.060 & 0.026 \\
 & 6 & 0.047 & 0.030 & 0.020 & 0.065 & 0.112 & 0.047 \\
 & 9 & 0.058 & 0.034 & 0.052 & 0.070 & 0.118 & 0.065 \\
 & 12 & 0.104 & 0.056 & 0.046 & 0.201 & 0.112 & 0.098 \\
 & 15 & 0.055 & 0.043 & 0.056 & 0.188 & 0.069 & 0.100 \\
\midrule
\multirow{6}{*}{LTC} & 1 & 0.011 & 0.009 & 0.006 & 0.007 & 0.017 & 0.013 \\
 & 3 & 0.012 & 0.007 & 0.006 & 0.011 & 0.008 & 0.006 \\
 & 6 & 0.011 & 0.007 & 0.006 & 0.003 & 0.008 & 0.006 \\
 & 9 & 0.008 & 0.015 & 0.007 & 0.004 & 0.008 & 0.011 \\
 & 12 & 0.004 & 0.010 & 0.005 & 0.004 & 0.004 & 0.004 \\
 & 15 & 0.006 & 0.007 & 0.006 & 0.003 & 0.002 & 0.009 \\
\midrule
\multirow{6}{*}{ETC} & 1 & 0.004 & 0.006 & 0.004 & 0.004 & 0.003 & 0.003 \\
 & 3 & 0.003 & 0.003 & 0.006 & 0.003 & 0.004 & 0.004 \\
 & 6 & 0.002 & 0.005 & 0.004 & 0.006 & 0.009 & 0.007 \\
 & 9 & 0.001 & 0.017 & 0.003 & 0.010 & 0.006 & 0.009 \\
 & 12 & 0.004 & 0.009 & 0.014 & 0.003 & 0.020 & 0.013 \\
 & 15 & 0.004 & 0.030 & 0.004 & 0.007 & 0.014 & 0.009 \\
\midrule
\multirow{6}{*}{ETH} & 1 & 0.004 & 0.003 & 0.001 & 0.001 & 0.001 & 0.001 \\
 & 3 & 0.002 & 0.003 & 0.004 & 0.002 & 0.004 & 0.007 \\
 & 6 & 0.006 & 0.011 & 0.015 & 0.017 & 0.007 & 0.027 \\
 & 9 & 0.012 & 0.011 & 0.017 & 0.014 & 0.009 & 0.008 \\
 & 12 & 0.006 & 0.007 & 0.003 & 0.011 & 0.023 & 0.015 \\
 & 15 & 0.001 & 0.011 & 0.036 & 0.005 & 0.012 & 0.013 \\
\midrule
\multirow{6}{*}{XRP} & 1 & 0.023 & 0.026 & 0.026 & 0.010 & 0.016 & 0.024 \\
 & 3 & 0.019 & 0.021 & 0.016 & 0.011 & 0.033 & 0.005 \\
 & 6 & 0.015 & 0.033 & 0.030 & 0.020 & 0.026 & 0.027 \\
 & 9 & 0.008 & 0.016 & 0.019 & 0.011 & 0.020 & 0.015 \\
 & 12 & 0.008 & 0.027 & 0.010 & 0.018 & 0.018 & 0.013 \\
 & 15 & 0.028 & 0.035 & 0.021 & 0.015 & 0.029 & 0.024 \\
\midrule
\multirow{6}{*}{BCH} & 1 & 0.002 & 0.002 & 0.004 & 0.004 & 0.005 & 0.002 \\
 & 3 & 0.002 & 0.003 & 0.003 & 0.002 & 0.004 & 0.006 \\
 & 6 & 0.004 & 0.006 & 0.005 & 0.002 & 0.002 & 0.005 \\
 & 9 & 0.005 & 0.006 & 0.015 & 0.006 & 0.003 & 0.005 \\
 & 12 & 0.006 & 0.008 & 0.005 & 0.005 & 0.009 & 0.009 \\
 & 15 & 0.004 & 0.008 & 0.005 & 0.005 & 0.009 & 0.010 \\
\bottomrule
\end{tabular}%
}
\end{table}

\begin{table}[H]
\centering
\caption{Mean RMSE}
\label{tab:mean-rmse}
\resizebox{\textwidth}{!}{%
\begin{tabular}{llrrrrrr}
\toprule
Asset & Steps & GRU & MoE\_GRU & KAMoE\_GRU & LSTM & MoE\_LSTM & KAMoE\_LSTM \\
\midrule
\multirow{6}{*}{BTC} & 1 & 0.055 & 0.055 & 0.055 & 0.055 & 0.056 & 0.056 \\
 & 3 & 0.062 & 0.062 & 0.062 & 0.066 & 0.064 & 0.063 \\
 & 6 & 0.068 & 0.067 & 0.065 & 0.078 & 0.071 & 0.068 \\
 & 9 & 0.072 & 0.068 & 0.069 & 0.077 & 0.072 & 0.074 \\
 & 12 & 0.074 & 0.072 & 0.070 & 0.085 & 0.075 & 0.073 \\
 & 15 & 0.072 & 0.070 & 0.072 & 0.078 & 0.076 & 0.075 \\
\midrule
\multirow{6}{*}{LTC} & 1 & 0.015 & 0.015 & 0.015 & 0.015 & 0.016 & 0.015 \\
 & 3 & 0.017 & 0.017 & 0.017 & 0.017 & 0.017 & 0.017 \\
 & 6 & 0.017 & 0.018 & 0.017 & 0.018 & 0.018 & 0.018 \\
 & 9 & 0.018 & 0.018 & 0.018 & 0.018 & 0.018 & 0.018 \\
 & 12 & 0.018 & 0.018 & 0.018 & 0.018 & 0.018 & 0.018 \\
 & 15 & 0.018 & 0.018 & 0.018 & 0.018 & 0.018 & 0.018 \\
\midrule
\multirow{6}{*}{ETC} & 1 & 0.030 & 0.030 & 0.030 & 0.030 & 0.030 & 0.030 \\
 & 3 & 0.032 & 0.032 & 0.032 & 0.032 & 0.032 & 0.032 \\
 & 6 & 0.033 & 0.033 & 0.033 & 0.033 & 0.033 & 0.033 \\
 & 9 & 0.034 & 0.035 & 0.034 & 0.034 & 0.034 & 0.034 \\
 & 12 & 0.034 & 0.036 & 0.035 & 0.034 & 0.036 & 0.035 \\
 & 15 & 0.035 & 0.036 & 0.035 & 0.035 & 0.035 & 0.035 \\
\midrule
\multirow{6}{*}{ETH} & 1 & 0.053 & 0.053 & 0.053 & 0.053 & 0.053 & 0.053 \\
 & 3 & 0.057 & 0.057 & 0.057 & 0.057 & 0.057 & 0.057 \\
 & 6 & 0.059 & 0.060 & 0.059 & 0.059 & 0.059 & 0.059 \\
 & 9 & 0.059 & 0.060 & 0.059 & 0.060 & 0.060 & 0.059 \\
 & 12 & 0.059 & 0.059 & 0.059 & 0.059 & 0.059 & 0.059 \\
 & 15 & 0.059 & 0.059 & 0.060 & 0.059 & 0.059 & 0.059 \\
\midrule
\multirow{6}{*}{XRP} & 1 & 0.009 & 0.009 & 0.009 & 0.009 & 0.009 & 0.009 \\
 & 3 & 0.009 & 0.009 & 0.009 & 0.009 & 0.009 & 0.009 \\
 & 6 & 0.010 & 0.010 & 0.010 & 0.010 & 0.010 & 0.010 \\
 & 9 & 0.010 & 0.010 & 0.010 & 0.010 & 0.010 & 0.010 \\
 & 12 & 0.010 & 0.010 & 0.010 & 0.010 & 0.010 & 0.010 \\
 & 15 & 0.010 & 0.011 & 0.010 & 0.010 & 0.011 & 0.011 \\
\midrule
\multirow{6}{*}{BCH} & 1 & 0.036 & 0.036 & 0.036 & 0.036 & 0.036 & 0.036 \\
 & 3 & 0.037 & 0.037 & 0.037 & 0.037 & 0.037 & 0.037 \\
 & 6 & 0.038 & 0.038 & 0.038 & 0.038 & 0.038 & 0.038 \\
 & 9 & 0.039 & 0.039 & 0.039 & 0.039 & 0.039 & 0.039 \\
 & 12 & 0.039 & 0.039 & 0.039 & 0.039 & 0.039 & 0.039 \\
 & 15 & 0.039 & 0.039 & 0.039 & 0.039 & 0.039 & 0.039 \\
\bottomrule
\end{tabular}%
}
\end{table}

\begin{table}[H]
\centering
\caption{Standard Deviation of RMSE ($\times 10^{-3}$)}
\label{tab:std-rmse}
\resizebox{\textwidth}{!}{%
\begin{tabular}{llrrrrrr}
\toprule
Asset & Steps & GRU & MoE\_GRU & KAMoE\_GRU & LSTM & MoE\_LSTM & KAMoE\_LSTM \\
\midrule
\multirow{6}{*}{BTC} & 1 & 0.163 & 0.129 & 0.213 & 0.323 & 0.161 & 0.521 \\
 & 3 & 0.283 & 0.208 & 0.347 & 1.877 & 2.232 & 1.021 \\
 & 6 & 1.674 & 1.088 & 0.746 & 2.104 & 3.837 & 1.684 \\
 & 9 & 1.977 & 1.204 & 1.835 & 2.163 & 3.884 & 2.135 \\
 & 12 & 3.377 & 1.920 & 1.616 & 5.708 & 3.550 & 3.239 \\
 & 15 & 1.907 & 1.523 & 1.939 & 5.881 & 2.232 & 3.223 \\
\midrule
\multirow{6}{*}{LTC} & 1 & 0.138 & 0.107 & 0.073 & 0.084 & 0.210 & 0.157 \\
 & 3 & 0.140 & 0.085 & 0.070 & 0.127 & 0.086 & 0.065 \\
 & 6 & 0.122 & 0.076 & 0.062 & 0.032 & 0.084 & 0.066 \\
 & 9 & 0.086 & 0.162 & 0.080 & 0.040 & 0.087 & 0.124 \\
 & 12 & 0.043 & 0.106 & 0.053 & 0.045 & 0.049 & 0.042 \\
 & 15 & 0.063 & 0.074 & 0.065 & 0.036 & 0.026 & 0.093 \\
\midrule
\multirow{6}{*}{ETC} & 1 & 0.112 & 0.170 & 0.131 & 0.108 & 0.095 & 0.104 \\
 & 3 & 0.094 & 0.091 & 0.162 & 0.085 & 0.112 & 0.122 \\
 & 6 & 0.069 & 0.145 & 0.114 & 0.169 & 0.258 & 0.200 \\
 & 9 & 0.027 & 0.461 & 0.081 & 0.294 & 0.164 & 0.242 \\
 & 12 & 0.118 & 0.248 & 0.407 & 0.102 & 0.551 & 0.365 \\
 & 15 & 0.118 & 0.813 & 0.111 & 0.203 & 0.381 & 0.259 \\
\midrule
\multirow{6}{*}{ETH} & 1 & 0.139 & 0.122 & 0.052 & 0.038 & 0.053 & 0.027 \\
 & 3 & 0.090 & 0.109 & 0.137 & 0.060 & 0.126 & 0.239 \\
 & 6 & 0.212 & 0.386 & 0.529 & 0.566 & 0.252 & 0.910 \\
 & 9 & 0.394 & 0.383 & 0.566 & 0.474 & 0.306 & 0.265 \\
 & 12 & 0.198 & 0.237 & 0.085 & 0.364 & 0.758 & 0.506 \\
 & 15 & 0.037 & 0.355 & 1.166 & 0.160 & 0.400 & 0.419 \\
\midrule
\multirow{6}{*}{XRP} & 1 & 0.175 & 0.195 & 0.194 & 0.076 & 0.120 & 0.182 \\
 & 3 & 0.127 & 0.144 & 0.107 & 0.075 & 0.225 & 0.033 \\
 & 6 & 0.098 & 0.214 & 0.202 & 0.135 & 0.173 & 0.180 \\
 & 9 & 0.057 & 0.099 & 0.122 & 0.074 & 0.131 & 0.100 \\
 & 12 & 0.056 & 0.174 & 0.071 & 0.118 & 0.118 & 0.087 \\
 & 15 & 0.184 & 0.221 & 0.133 & 0.102 & 0.190 & 0.151 \\
\midrule
\multirow{6}{*}{BCH} & 1 & 0.059 & 0.055 & 0.089 & 0.087 & 0.133 & 0.050 \\
 & 3 & 0.052 & 0.070 & 0.067 & 0.048 & 0.099 & 0.142 \\
 & 6 & 0.098 & 0.136 & 0.118 & 0.054 & 0.051 & 0.101 \\
 & 9 & 0.098 & 0.122 & 0.317 & 0.134 & 0.067 & 0.101 \\
 & 12 & 0.129 & 0.168 & 0.095 & 0.115 & 0.181 & 0.181 \\
 & 15 & 0.080 & 0.169 & 0.103 & 0.102 & 0.177 & 0.211 \\
\bottomrule
\end{tabular}%
}
\end{table}

\begin{table}[H]
\centering
\caption{Mean Training Time (seconds)}
\label{tab:mean-training-time}
\resizebox{\textwidth}{!}{%
\begin{tabular}{llrrrrrr}
\toprule
Asset & Steps & GRU & MoE\_GRU & KAMoE\_GRU & LSTM & MoE\_LSTM & KAMoE\_LSTM \\
\midrule
\multirow{6}{*}{BTC} & 1 & 27.55 & 55.51 & 71.71 & 20.79 & 51.98 & 55.39 \\
 & 3 & 34.80 & 70.15 & 77.08 & 23.78 & 55.29 & 59.36 \\
 & 6 & 38.62 & 68.10 & 74.61 & 23.16 & 58.57 & 59.50 \\
 & 9 & 35.22 & 64.98 & 72.55 & 22.97 & 57.84 & 62.81 \\
 & 12 & 44.37 & 86.27 & 85.02 & 27.93 & 71.99 & 71.55 \\
 & 15 & 45.77 & 93.70 & 94.44 & 28.76 & 76.41 & 90.06 \\
\midrule
\multirow{6}{*}{LTC} & 1 & 20.44 & 50.54 & 54.58 & 17.53 & 48.08 & 52.44 \\
 & 3 & 21.36 & 52.24 & 54.28 & 18.31 & 47.96 & 52.89 \\
 & 6 & 21.64 & 51.62 & 54.67 & 18.17 & 47.89 & 54.97 \\
 & 9 & 21.40 & 52.53 & 54.23 & 17.14 & 49.62 & 51.50 \\
 & 12 & 26.99 & 62.88 & 63.69 & 21.40 & 56.58 & 61.42 \\
 & 15 & 30.88 & 70.07 & 72.75 & 24.46 & 65.21 & 68.95 \\
\midrule
\multirow{6}{*}{ETC} & 1 & 24.06 & 55.19 & 60.91 & 17.58 & 51.35 & 56.75 \\
 & 3 & 25.42 & 55.73 & 65.01 & 19.40 & 49.78 & 55.99 \\
 & 6 & 24.08 & 59.57 & 63.16 & 22.17 & 51.09 & 59.20 \\
 & 9 & 23.45 & 55.24 & 61.17 & 21.52 & 52.80 & 59.61 \\
 & 12 & 27.65 & 62.03 & 68.12 & 25.22 & 64.16 & 70.16 \\
 & 15 & 36.36 & 74.78 & 78.54 & 28.27 & 68.30 & 73.43 \\
\midrule
\multirow{6}{*}{ETH} & 1 & 24.34 & 61.50 & 59.85 & 21.75 & 54.38 & 57.76 \\
 & 3 & 30.07 & 63.31 & 65.02 & 23.84 & 57.06 & 70.27 \\
 & 6 & 27.42 & 83.99 & 71.75 & 28.45 & 60.38 & 72.27 \\
 & 9 & 29.85 & 84.33 & 69.60 & 25.02 & 60.18 & 61.39 \\
 & 12 & 38.23 & 82.89 & 70.96 & 29.73 & 69.93 & 67.31 \\
 & 15 & 48.82 & 102.10 & 116.28 & 32.61 & 75.71 & 86.59 \\
\midrule
\multirow{6}{*}{XRP} & 1 & 21.94 & 51.75 & 55.01 & 18.19 & 47.65 & 53.61 \\
 & 3 & 21.45 & 53.40 & 55.08 & 18.32 & 48.91 & 53.82 \\
 & 6 & 21.98 & 53.09 & 55.63 & 17.80 & 49.39 & 53.00 \\
 & 9 & 21.29 & 52.16 & 55.32 & 18.26 & 48.75 & 52.32 \\
 & 12 & 26.66 & 59.89 & 63.57 & 22.08 & 58.58 & 61.34 \\
 & 15 & 30.59 & 73.11 & 74.55 & 24.61 & 66.44 & 69.45 \\
\midrule
\multirow{6}{*}{BCH} & 1 & 21.67 & 59.54 & 63.97 & 17.39 & 51.32 & 54.11 \\
 & 3 & 25.39 & 57.43 & 62.10 & 18.76 & 55.65 & 56.19 \\
 & 6 & 23.00 & 53.71 & 57.87 & 19.01 & 53.66 & 54.61 \\
 & 9 & 21.06 & 54.96 & 60.05 & 18.37 & 50.70 & 54.64 \\
 & 12 & 26.37 & 63.25 & 64.94 & 22.00 & 60.30 & 62.66 \\
 & 15 & 31.90 & 73.16 & 77.60 & 24.98 & 66.92 & 70.82 \\
\bottomrule
\end{tabular}%
}
\end{table}

\begin{table}[H]
\centering
\caption{Standard Deviation of Training Time (seconds)}
\label{tab:std-training-time}
\resizebox{\textwidth}{!}{%
\begin{tabular}{llrrrrrr}
\toprule
Asset & Steps & GRU & MoE\_GRU & KAMoE\_GRU & LSTM & MoE\_LSTM & KAMoE\_LSTM \\
\midrule
\multirow{6}{*}{BTC} & 1 & 3.87 & 3.01 & 7.41 & 2.04 & 2.80 & 5.13 \\
 & 3 & 4.00 & 4.10 & 9.39 & 2.27 & 6.01 & 3.46 \\
 & 6 & 3.93 & 9.09 & 10.20 & 2.24 & 5.40 & 5.83 \\
 & 9 & 3.51 & 4.04 & 7.66 & 1.37 & 6.53 & 7.51 \\
& 12 & 8.86 & 14.73 & 9.19 & 2.71 & 10.77 & 5.45 \\
 & 15 & 4.79 & 13.13 & 10.39 & 3.11 & 6.39 & 11.62 \\
\midrule
\multirow{6}{*}{LTC} & 1 & 0.49 & 1.24 & 2.19 & 1.13 & 2.42 & 2.66 \\
 & 3 & 1.29 & 3.65 & 2.32 & 0.79 & 1.75 & 2.58 \\
 & 6 & 0.64 & 2.22 & 1.55 & 0.98 & 1.22 & 4.37 \\
 & 9 & 1.57 & 3.66 & 1.81 & 0.13 & 3.00 & 1.89 \\
 & 12 & 3.79 & 4.64 & 1.64 & 1.73 & 1.81 & 2.69 \\
 & 15 & 1.04 & 3.36 & 2.02 & 0.56 & 2.21 & 1.74 \\
\midrule
\multirow{6}{*}{ETC} & 1 & 2.99 & 3.56 & 7.21 & 1.26 & 2.42 & 5.91 \\
 & 3 & 3.63 & 4.00 & 8.66 & 1.64 & 2.69 & 3.96 \\
 & 6 & 2.09 & 2.05 & 6.27 & 1.89 & 4.80 & 5.56 \\
 & 9 & 2.30 & 2.33 & 3.43 & 3.10 & 5.86 & 6.33 \\
 & 12 & 1.97 & 0.84 & 4.93 & 3.25 & 6.04 & 6.78 \\
 & 15 & 4.41 & 5.73 & 6.32 & 3.06 & 5.02 & 3.68 \\
\midrule
\multirow{6}{*}{ETH} & 1 & 3.94 & 4.38 & 6.62 & 2.14 & 3.45 & 4.08 \\
 & 3 & 4.35 & 8.34 & 3.13 & 1.79 & 6.15 & 6.20 \\
 & 6 & 4.71 & 5.53 & 5.09 & 2.81 & 2.92 & 7.43 \\
 & 9 & 5.96 & 9.74 & 9.77 & 2.38 & 3.64 & 4.85 \\
 & 12 & 6.33 & 7.85 & 5.48 & 3.50 & 6.85 & 3.59 \\
 & 15 & 3.69 & 11.27 & 12.16 & 2.77 & 5.49 & 5.87 \\
\midrule
\multirow{6}{*}{XRP} & 1 & 1.76 & 1.81 & 2.27 & 1.13 & 1.74 & 2.12 \\
 & 3 & 0.29 & 2.91 & 2.21 & 1.11 & 1.72 & 4.38 \\
 & 6 & 1.06 & 3.98 & 2.01 & 0.52 & 2.77 & 3.53 \\
 & 9 & 0.55 & 1.11 & 2.16 & 1.43 & 1.33 & 2.47 \\
 & 12 & 1.60 & 1.22 & 1.54 & 0.85 & 4.50 & 2.06 \\
 & 15 & 1.10 & 8.40 & 2.15 & 0.41 & 3.42 & 0.58 \\
\midrule
\multirow{6}{*}{BCH} & 1 & 0.61 & 5.45 & 7.53 & 1.32 & 2.00 & 3.05 \\
 & 3 & 4.17 & 2.12 & 3.64 & 1.66 & 7.21 & 3.89 \\
 & 6 & 1.70 & 2.52 & 3.11 & 1.79 & 3.77 & 3.22 \\
 & 9 & 0.33 & 3.34 & 3.05 & 1.54 & 2.89 & 3.32 \\
 & 12 & 1.07 & 2.35 & 2.15 & 0.88 & 2.76 & 1.32 \\
 & 15 & 1.34 & 6.48 & 4.19 & 1.10 & 3.90 & 2.45 \\
\bottomrule
\end{tabular}%
}
\end{table}

\section{Additional Results Task 2}

\begin{table}[h]
\centering
\caption{KANLinear: Standard Deviation of R2 Values}
\begin{tabular}{|c|cccc|cccc|cccc|}
\hline
\multirow{2}{*}{\shortstack{Hidden\\Units}} & \multicolumn{4}{c|}{KAMoE} & \multicolumn{4}{c|}{MoE} & \multicolumn{4}{c|}{Standard} \\
 & \multicolumn{12}{c|}{Number of Layers} \\
 & 1 & 2 & 3 & 4 & 1 & 2 & 3 & 4 & 1 & 2 & 3 & 4 \\
\hline
5 & 0.0061 & 0.0081 & 0.0057 & 0.0105 & 0.0068 & 0.0034 & 0.0044 & 0.0095 & 0.0016 & 0.0119 & 0.0124 & 0.0124 \\
10 & 0.0099 & 0.0054 & 0.0089 & 0.0055 & 0.0059 & 0.0031 & 0.0106 & 0.0031 & 0.0007 & 0.0067 & 0.0050 & 0.0100 \\
25 & 0.0039 & 0.0049 & 0.0074 & 0.0079 & 0.0033 & 0.0053 & 0.0045 & 0.0090 & 0.0008 & 0.0035 & 0.0022 & 0.0047 \\
50 & 0.0062 & 0.0033 & 0.0085 & 0.0113 & 0.0040 & 0.0020 & 0.0071 & 0.0083 & 0.0007 & 0.0042 & 0.0089 & 0.0058 \\
100 & 0.0043 & 0.0028 & 0.0057 & 0.0072 & 0.0030 & 0.0023 & 0.0123 & 0.0036 & 0.0007 & 0.0032 & 0.0040 & 0.0058 \\
200 & 0.0027 & 0.0041 & 0.0038 & 0.0066 & 0.0029 & 0.0043 & 0.0056 & 0.0047 & 0.0004 & 0.0044 & 0.0040 & 0.0041 \\
400 & 0.0033 & 0.0031 & 0.0058 & 0.0058 & 0.0031 & 0.0066 & 0.0039 & 0.0015 & 0.0004 & 0.0023 & 0.0041 & 0.0021 \\
800 & 0.0049 & 0.0165 & 0.0180 & 0.0082 & 0.0055 & 0.0072 & 0.0029 & 0.0014 & 0.0001 & 0.0016 & 0.0020 & 0.0032 \\
\hline
\end{tabular}
\end{table}

\begin{table}[h]
\centering
\caption{MLP: Standard Deviation of R2 Values}
\begin{tabular}{|c|cccc|cccc|cccc|}
\hline
\multirow{2}{*}{\shortstack{Hidden\\Units}} & \multicolumn{4}{c|}{KAMoE} & \multicolumn{4}{c|}{MoE} & \multicolumn{4}{c|}{Standard} \\
 & \multicolumn{12}{c|}{Number of Layers} \\
 & 1 & 2 & 3 & 4 & 1 & 2 & 3 & 4 & 1 & 2 & 3 & 4 \\
\hline
5 & 0.0121 & 0.0126 & 0.0056 & 0.0053 & 0.0100 & 0.0079 & 0.0078 & 0.0090 & 0.0259 & 0.0135 & 0.0075 & 0.0075 \\
10 & 0.0036 & 0.0048 & 0.0023 & 0.0053 & 0.0050 & 0.0079 & 0.0046 & 0.0045 & 0.0096 & 0.0053 & 0.0047 & 0.0041 \\
25 & 0.0069 & 0.0043 & 0.0037 & 0.0033 & 0.0071 & 0.0043 & 0.0052 & 0.0035 & 0.0029 & 0.0024 & 0.0024 & 0.0037 \\
50 & 0.0020 & 0.0033 & 0.0039 & 0.0021 & 0.0033 & 0.0022 & 0.0030 & 0.0023 & 0.0009 & 0.0029 & 0.0041 & 0.0039 \\
100 & 0.0047 & 0.0056 & 0.0020 & 0.0037 & 0.0042 & 0.0021 & 0.0043 & 0.0029 & 0.0022 & 0.0007 & 0.0036 & 0.0011 \\
200 & 0.0048 & 0.0034 & 0.0019 & 0.0023 & 0.0037 & 0.0044 & 0.0017 & 0.0025 & 0.0023 & 0.0032 & 0.0009 & 0.0032 \\
400 & 0.0019 & 0.0028 & 0.0021 & 0.0022 & 0.0043 & 0.0038 & 0.0017 & 0.0037 & 0.0025 & 0.0015 & 0.0019 & 0.0016 \\
800 & 0.0023 & 0.0033 & 0.0042 & 0.0023 & 0.0015 & 0.0008 & 0.0017 & 0.0020 & 0.0022 & 0.0011 & 0.0012 & 0.0025 \\
\hline
\end{tabular}
\end{table}

\begin{table}[h]
\centering
\caption{KANLinear: Mean MSE Values}
\begin{tabular}{|c|cccc|cccc|cccc|}
\hline
\multirow{2}{*}{\shortstack{Hidden\\Units}} & \multicolumn{4}{c|}{KAMoE} & \multicolumn{4}{c|}{MoE} & \multicolumn{4}{c|}{Standard} \\
 & \multicolumn{12}{c|}{Number of Layers} \\
 & 1 & 2 & 3 & 4 & 1 & 2 & 3 & 4 & 1 & 2 & 3 & 4 \\
\hline
5 & 0.305 & 0.312 & 0.307 & 0.321 & 0.317 & 0.303 & 0.319 & 0.326 & 0.497 & 0.405 & 0.408 & 0.432 \\
10 & 0.308 & 0.292 & 0.314 & 0.324 & 0.304 & 0.300 & 0.317 & 0.327 & 0.496 & 0.387 & 0.396 & 0.416 \\
25 & 0.304 & 0.302 & 0.323 & 0.335 & 0.299 & 0.306 & 0.327 & 0.346 & 0.496 & 0.395 & 0.401 & 0.417 \\
50 & 0.294 & 0.308 & 0.327 & 0.334 & 0.294 & 0.313 & 0.335 & 0.332 & 0.496 & 0.394 & 0.411 & 0.414 \\
100 & 0.285 & 0.308 & 0.315 & 0.320 & 0.288 & 0.312 & 0.330 & 0.317 & 0.496 & 0.402 & 0.414 & 0.413 \\
200 & 0.284 & 0.311 & 0.308 & 0.302 & 0.283 & 0.310 & 0.304 & 0.308 & 0.496 & 0.403 & 0.408 & 0.396 \\
400 & 0.286 & 0.330 & 0.304 & 0.299 & 0.281 & 0.333 & 0.304 & 0.300 & 0.496 & 0.408 & 0.395 & 0.386 \\
800 & 0.286 & 0.340 & 0.308 & 0.298 & 0.277 & 0.338 & 0.303 & 0.297 & 0.495 & 0.399 & 0.389 & 0.383 \\
\hline
\end{tabular}
\end{table}

\begin{table}[h]
\centering
\caption{MLP: Mean MSE Values}
\begin{tabular}{|c|cccc|cccc|cccc|}
\hline
\multirow{2}{*}{\shortstack{Hidden\\Units}} & \multicolumn{4}{c|}{KAMoE} & \multicolumn{4}{c|}{MoE} & \multicolumn{4}{c|}{Standard} \\
 & \multicolumn{12}{c|}{Number of Layers} \\
 & 1 & 2 & 3 & 4 & 1 & 2 & 3 & 4 & 1 & 2 & 3 & 4 \\
\hline
5 & 0.324 & 0.310 & 0.303 & 0.303 & 0.321 & 0.312 & 0.305 & 0.317 & 0.399 & 0.357 & 0.359 & 0.358 \\
10 & 0.305 & 0.295 & 0.299 & 0.296 & 0.310 & 0.303 & 0.293 & 0.289 & 0.350 & 0.321 & 0.316 & 0.309 \\
25 & 0.295 & 0.292 & 0.285 & 0.285 & 0.295 & 0.285 & 0.287 & 0.284 & 0.327 & 0.297 & 0.289 & 0.282 \\
50 & 0.292 & 0.280 & 0.278 & 0.278 & 0.296 & 0.282 & 0.279 & 0.277 & 0.319 & 0.287 & 0.282 & 0.280 \\
100 & 0.287 & 0.274 & 0.269 & 0.273 & 0.286 & 0.277 & 0.272 & 0.272 & 0.312 & 0.275 & 0.270 & 0.271 \\
200 & 0.284 & 0.272 & 0.267 & 0.267 & 0.283 & 0.273 & 0.270 & 0.266 & 0.310 & 0.270 & 0.269 & 0.268 \\
400 & 0.282 & 0.268 & 0.263 & 0.266 & 0.281 & 0.265 & 0.265 & 0.269 & 0.308 & 0.261 & 0.262 & 0.266 \\
800 & 0.278 & 0.265 & 0.261 & 0.265 & 0.280 & 0.266 & 0.264 & 0.265 & 0.299 & 0.258 & 0.262 & 0.261 \\
\hline
\end{tabular}
\end{table}

\begin{table}[h]
\centering
\caption{KANLinear: Standard Deviation of MSE Values}
\begin{tabular}{|c|cccc|cccc|cccc|}
\hline
\multirow{2}{*}{\shortstack{Hidden\\Units}} & \multicolumn{4}{c|}{KAMoE} & \multicolumn{4}{c|}{MoE} & \multicolumn{4}{c|}{Standard} \\
 & \multicolumn{12}{c|}{Number of Layers} \\
 & 1 & 2 & 3 & 4 & 1 & 2 & 3 & 4 & 1 & 2 & 3 & 4 \\
\hline
5 & 0.0080 & 0.0106 & 0.0075 & 0.0137 & 0.0089 & 0.0044 & 0.0057 & 0.0125 & 0.0021 & 0.0156 & 0.0162 & 0.0162 \\
10 & 0.0129 & 0.0070 & 0.0117 & 0.0072 & 0.0077 & 0.0041 & 0.0139 & 0.0041 & 0.0009 & 0.0088 & 0.0066 & 0.0131 \\
25 & 0.0051 & 0.0065 & 0.0097 & 0.0104 & 0.0044 & 0.0070 & 0.0059 & 0.0117 & 0.0010 & 0.0045 & 0.0029 & 0.0061 \\
50 & 0.0081 & 0.0044 & 0.0111 & 0.0149 & 0.0053 & 0.0026 & 0.0092 & 0.0109 & 0.0010 & 0.0054 & 0.0117 & 0.0076 \\
100 & 0.0056 & 0.0037 & 0.0074 & 0.0094 & 0.0039 & 0.0030 & 0.0161 & 0.0048 & 0.0010 & 0.0042 & 0.0052 & 0.0076 \\
200 & 0.0035 & 0.0053 & 0.0049 & 0.0087 & 0.0039 & 0.0056 & 0.0073 & 0.0061 & 0.0005 & 0.0057 & 0.0052 & 0.0053 \\
400 & 0.0043 & 0.0041 & 0.0076 & 0.0077 & 0.0040 & 0.0087 & 0.0051 & 0.0020 & 0.0005 & 0.0030 & 0.0054 & 0.0028 \\
800 & 0.0064 & 0.0216 & 0.0236 & 0.0108 & 0.0072 & 0.0095 & 0.0038 & 0.0018 & 0.0001 & 0.0021 & 0.0026 & 0.0041 \\
\hline
\end{tabular}
\end{table}

\begin{table}[h]
\centering
\caption{MLP: Standard Deviation of MSE Values}
\begin{tabular}{|c|cccc|cccc|cccc|}
\hline
\multirow{2}{*}{\shortstack{Hidden\\Units}} & \multicolumn{4}{c|}{KAMoE} & \multicolumn{4}{c|}{MoE} & \multicolumn{4}{c|}{Standard} \\
 & \multicolumn{12}{c|}{Number of Layers} \\
 & 1 & 2 & 3 & 4 & 1 & 2 & 3 & 4 & 1 & 2 & 3 & 4 \\
\hline
5 & 0.0159 & 0.0166 & 0.0074 & 0.0069 & 0.0131 & 0.0104 & 0.0102 & 0.0118 & 0.0339 & 0.0177 & 0.0098 & 0.0098 \\
10 & 0.0047 & 0.0062 & 0.0031 & 0.0069 & 0.0066 & 0.0104 & 0.0060 & 0.0060 & 0.0126 & 0.0069 & 0.0061 & 0.0054 \\
25 & 0.0091 & 0.0056 & 0.0049 & 0.0043 & 0.0093 & 0.0056 & 0.0068 & 0.0046 & 0.0038 & 0.0031 & 0.0032 & 0.0049 \\
50 & 0.0026 & 0.0043 & 0.0050 & 0.0027 & 0.0043 & 0.0029 & 0.0039 & 0.0030 & 0.0012 & 0.0038 & 0.0054 & 0.0051 \\
100 & 0.0062 & 0.0073 & 0.0026 & 0.0048 & 0.0055 & 0.0027 & 0.0056 & 0.0038 & 0.0029 & 0.0009 & 0.0048 & 0.0014 \\
200 & 0.0063 & 0.0045 & 0.0025 & 0.0030 & 0.0049 & 0.0058 & 0.0022 & 0.0033 & 0.0030 & 0.0042 & 0.0011 & 0.0042 \\
400 & 0.0025 & 0.0037 & 0.0027 & 0.0029 & 0.0056 & 0.0050 & 0.0022 & 0.0049 & 0.0032 & 0.0019 & 0.0025 & 0.0021 \\
800 & 0.0030 & 0.0044 & 0.0055 & 0.0030 & 0.0020 & 0.0010 & 0.0022 & 0.0027 & 0.0028 & 0.0015 & 0.0016 & 0.0032 \\
\hline
\end{tabular}
\end{table}

\begin{table}[h]
\centering
\caption{KANLinear: Mean Time Values (in seconds)}
\begin{tabular}{|c|cccc|cccc|cccc|}
\hline
\multirow{2}{*}{\shortstack{Hidden\\Units}} & \multicolumn{4}{c|}{KAMoE} & \multicolumn{4}{c|}{MoE} & \multicolumn{4}{c|}{Standard} \\
 & \multicolumn{12}{c|}{Number of Layers} \\
 & 1 & 2 & 3 & 4 & 1 & 2 & 3 & 4 & 1 & 2 & 3 & 4 \\
\hline
5 & 55.32 & 100.64 & 118.63 & 138.04 & 66.54 & 84.89 & 96.58 & 108.80 & 14.45 & 23.73 & 20.42 & 38.63 \\
10 & 56.62 & 67.07 & 83.68 & 103.41 & 47.33 & 66.07 & 74.86 & 85.02 & 10.39 & 17.15 & 16.92 & 19.61 \\
25 & 41.72 & 54.45 & 65.45 & 89.20 & 32.87 & 45.56 & 52.63 & 72.48 & 9.14 & 12.54 & 12.36 & 14.83 \\
50 & 44.43 & 50.70 & 65.12 & 83.17 & 33.07 & 42.58 & 53.72 & 67.16 & 7.73 & 10.46 & 11.12 & 13.38 \\
100 & 40.57 & 49.21 & 61.75 & 80.74 & 33.08 & 38.73 & 54.64 & 68.04 & 7.95 & 10.38 & 11.24 & 13.79 \\
200 & 45.69 & 48.19 & 65.67 & 82.30 & 34.24 & 42.10 & 53.80 & 71.12 & 9.04 & 10.34 & 11.73 & 14.67 \\
400 & 43.84 & 50.22 & 70.00 & 89.24 & 37.25 & 42.66 & 60.32 & 72.29 & 8.39 & 10.34 & 12.64 & 15.71 \\
800 & 47.18 & 54.07 & 82.58 & 100.57 & 33.36 & 48.52 & 69.21 & 79.60 & 8.99 & 10.49 & 15.02 & 18.15 \\
\hline
\end{tabular}
\end{table}

\begin{table}[h]
\centering
\caption{MLP: Mean Time Values (in seconds)}
\begin{tabular}{|c|cccc|cccc|cccc|}
\hline
\multirow{2}{*}{\shortstack{Hidden\\Units}} & \multicolumn{4}{c|}{KAMoE} & \multicolumn{4}{c|}{MoE} & \multicolumn{4}{c|}{Standard} \\
 & \multicolumn{12}{c|}{Number of Layers} \\
 & 1 & 2 & 3 & 4 & 1 & 2 & 3 & 4 & 1 & 2 & 3 & 4 \\
\hline
5 & 50.61 & 79.67 & 75.23 & 80.75 & 28.42 & 43.67 & 53.50 & 67.08 & 17.59 & 22.59 & 18.95 & 14.87 \\
10 & 36.79 & 44.95 & 72.77 & 84.54 & 43.09 & 41.17 & 65.27 & 71.60 & 17.18 & 20.15 & 16.13 & 22.19 \\
25 & 35.26 & 42.62 & 59.26 & 75.78 & 29.39 & 37.58 & 40.08 & 53.66 & 21.15 & 21.98 & 15.06 & 15.33 \\
50 & 30.36 & 37.90 & 55.01 & 65.47 & 20.66 & 34.10 & 38.92 & 48.91 & 19.03 & 16.77 & 12.98 & 15.20 \\
100 & 29.63 & 45.48 & 55.32 & 69.46 & 23.34 & 33.94 & 39.38 & 45.23 & 16.73 & 15.48 & 10.36 & 10.16 \\
200 & 24.52 & 47.69 & 55.19 & 62.29 & 24.27 & 30.05 & 39.53 & 46.21 & 14.10 & 19.50 & 12.22 & 10.16 \\
400 & 28.09 & 46.29 & 52.46 & 65.45 & 25.57 & 34.68 & 35.86 & 44.78 & 12.58 & 15.60 & 12.64 & 12.32 \\
800 & 38.75 & 53.89 & 52.69 & 59.20 & 22.93 & 32.79 & 39.06 & 42.71 & 11.56 & 14.13 & 9.55 & 8.93 \\
\hline
\end{tabular}
\end{table}

\begin{table}[h]
\centering
\caption{KANLinear: Mean Number of Parameters}
\begin{tabular}{|c|cccc|cccc|cccc|}
\hline
\multirow{2}{*}{\shortstack{Hidden\\Units}} & \multicolumn{4}{c|}{KAMoE} & \multicolumn{4}{c|}{MoE} & \multicolumn{4}{c|}{Standard} \\
 & \multicolumn{12}{c|}{Number of Layers} \\
 & 1 & 2 & 3 & 4 & 1 & 2 & 3 & 4 & 1 & 2 & 3 & 4 \\
\hline
5 & 1228 & 2099 & 2970 & 3841 & 1070 & 1783 & 2496 & 3209 & 317 & 517 & 717 & 917 \\
10 & 2253 & 4949 & 7645 & 10341 & 2005 & 4453 & 6901 & 9349 & 607 & 1347 & 2087 & 2827 \\
25 & 5328 & 20099 & 34870 & 49641 & 4810 & 19063 & 33316 & 47569 & 1477 & 5937 & 10397 & 14857 \\
50 & 10453 & 67349 & 124245 & 181141 & 9485 & 65413 & 121341 & 177269 & 2927 & 20587 & 38247 & 55907 \\
100 & 20703 & 244349 & 467995 & 691641 & 18835 & 240613 & 462391 & 684169 & 5827 & 76137 & 146447 & 216757 \\
200 & 41203 & 928349 & 1815495 & 2702641 & 37535 & 921013 & 1804491 & 2687969 & 11627 & 292237 & 572847 & 853457 \\
400 & 82203 & 3616349 & 7150495 & 10684641 & 74935 & 3601813 & 7128691 & 10655569 & 23227 & 1144437 & 2265647 & 3386857 \\
800 & 164203 & 14272349 & 28380495 & 42488641 & 149735 & 14243413 & 28337091 & 42430769 & 46427 & 4528837 & 9011247 & 13493657 \\
\hline
\end{tabular}
\end{table}

\begin{table}[h]
\centering
\caption{MLP: Mean Number of Parameters}
\begin{tabular}{|c|cccc|cccc|cccc|}
\hline
\multirow{2}{*}{\shortstack{Hidden\\Units}} & \multicolumn{4}{c|}{KAMoE} & \multicolumn{4}{c|}{MoE} & \multicolumn{4}{c|}{Standard} \\
 & \multicolumn{12}{c|}{Number of Layers} \\
 & 1 & 2 & 3 & 4 & 1 & 2 & 3 & 4 & 1 & 2 & 3 & 4 \\
\hline
5 & 430 & 791 & 1152 & 1513 & 272 & 475 & 678 & 881 & 51 & 81 & 111 & 141 \\
10 & 735 & 1541 & 2347 & 3153 & 487 & 1045 & 1603 & 2161 & 101 & 211 & 321 & 431 \\
25 & 1650 & 4991 & 8332 & 11673 & 1132 & 3955 & 6778 & 9601 & 251 & 901 & 1551 & 2201 \\
50 & 3175 & 14741 & 26307 & 37873 & 2207 & 12805 & 23403 & 34001 & 501 & 3051 & 5601 & 8151 \\
100 & 6225 & 49241 & 92257 & 135273 & 4357 & 45505 & 86653 & 127801 & 1001 & 11101 & 21201 & 31301 \\
200 & 12325 & 178241 & 344157 & 510073 & 8657 & 170905 & 333153 & 495401 & 2001 & 42201 & 82401 & 122601 \\
400 & 24525 & 676241 & 1327957 & 1979673 & 17257 & 661705 & 1306153 & 1950601 & 4001 & 164401 & 324801 & 485201 \\
800 & 48925 & 2632241 & 5215557 & 7798873 & 34457 & 2603305 & 5172153 & 7741001 & 8001 & 648801 & 1289601 & 1930401 \\
\hline
\end{tabular}
\end{table}

\end{document}